\pdfoutput=1

\documentclass[11pt]{article}
\usepackage{CJKutf8}

\usepackage[preprint]{acl}

\usepackage{times}
\usepackage{latexsym}
\usepackage{float}

\usepackage[T1]{fontenc}

\usepackage[utf8]{inputenc}
\usepackage{subfig}

\usepackage{microtype}

\usepackage{inconsolata}

\usepackage{graphicx}
\usepackage{verbatim}

\usepackage{csquotes}

\title{What talking you?: Translating Code-Mixed Messaging Texts to English}

\author{Lynnette Hui Xian Ng\textsuperscript{*} \\
  Carnegie Mellon University \\
  \texttt{lynnetteng@cmu.edu} \\\And
  Luo Qi Chan\textsuperscript{*} \\
  Carnegie Mellon University \\
  \texttt{luoqic@andrew.cmu.edu} \\
  }

\begin{document}
\maketitle
\def\thefootnote{*}\footnotetext{These authors contributed equally to this work.}\def\thefootnote{\arabic{footnote}}

\begin{abstract}
Translation of code-mixed texts to formal English allow a wider audience to understand these code-mixed languages, and facilitate downstream analysis applications such as sentiment analysis. In this work, we look at translating Singlish, which is colloquial Singaporean English, to formal standard English. Singlish is formed through the code-mixing of multiple Asian languages and dialects. We analysed the presence of other Asian languages and variants which can facilitate translation. Our dataset is short message texts, written as informal communication between Singlish speakers. We use a multi-step prompting scheme on five Large Language Models (LLMs) for language detection and translation. Our analysis show that LLMs do not perform well in this task, and we describe the challenges involved in translation of code-mixed languages.
We also release our dataset in this link\footnote{\url{https://github.com/luoqichan/singlish}}.
\end{abstract}

\section{Introduction}
Code-mixing occurs when more than one language is used in a given sentence. This phenomenon is prevalent in multicultural and multilingual societies such as Singapore, India, Malaysia and so forth. Code-mixing is widely used in informal settings, such as text messages and social media communication \cite{nayak2022l3cube,jayanthi2021codemixednlp}. Developing tools to computationally translate code-mixed languages contribute to the comprehension of these mixed texts, particularly in the area of social understanding, and improving Artificial Intelligent (AI) chat assistants.

Pre-trained multilingual language models such as mBERT and XLMR have been used to tackle the code-mixing problem, as they have been trained on over 100 languages \cite{santy2021bertologicomix}. \citet{jayanthi2021codemixednlp} created transformer-based models for seven tasks for analysing Hindi-English sentences, which includes sentiment classification, Language Detection and machine translation. The advent of Large Language Models (LLMs) and their exceptional performance in machine translation tasks presents an opportunity to perform tasks on code-mixed languages. \citet{gupta2024code} used k-shot prompting and rule-based prompting techniques to create a code-mixed dataset spanning five language pairs, and \citet{yadav2024code} harnessed LLMs to evaluate sentiments of code-mixed sentences in five language pairs.

Studies of code-mixed languages mostly work with a pair of languages, typically a native language and English. Here, we build on past work and tackle Singlish. Singlish is a code-mixed language that comprise several languages and variants in a single sentence. It is a rich language from Singapore that incorporates many South Asian languages and dialects, such as Chinese, Cantonese, Malay, Hindi and so forth, within the same sentence.

Translating Singlish to English increases the accessibility and understanding of the language. With a dataset derived from text messages written in Singlish, we perform experiments on two tasks: Language Detection and Sentence Translation. Using LLMs as our language inference models, we find that it is not easy to predict the language present in the texts and effectively translate the texts. 

\section{Literature Review}
Machine translation is an active research area in Natural Language Processing (NLP) with a long string of works. These involves Statistical Machine Translation (SMT) \cite{lopez2008statistical,wang2017neural}, which most models were defined with a log-linear framework \cite{och2002discriminative}, that led to Neural Machine Translation methods \cite{sharma2021review}. However, SMT produced adequate but non-fluent translations and NMT produced fluent but inadequate translations
\cite{wang2017neural,tu2017context}. Therefore, approaches to improve translation capabilities involved a multi-task NMT model \cite{tebbifakhr2020automatic}, an alignment-based models \cite{bahdanau2014neural}, and so forth. Soon, Deep Learning Techniques were introduced in machine translation, which provided better accuracies with attention-based, transformer-based, and encoder based machine learning networks \cite{zhang2015deep,yang2020survey}.

Recently, Large Language Models have been employed in the machine translation task \cite{chitale2024empirical,jiao2023chatgpt,liao2024ikun}, due to its remarkable ability to perform a wide variety of tasks. \citet{jiao2023chatgpt} provided a preliminary investigation for the use of GPT-4 for its multilingual translation capabilities, revealing that it performs well on high-resource European languages, but lags behind on Reddit posts. \citet{liao2024ikun} further evaluated the capability of LLMs for multilingual machine translation, as an alternative to training bilingual translation systems that work for translation of only two languages from scratch. 
For low-resource languages that do not have much training data data, which code-mixed languages fall under, \citet{enis2024llm} used Claude 3 Opus outperforms Google Translate in Maltese-English sentence pairs. Translation in low-resource language can also be approached through identifying parallel data and creation of task diversity \cite{iyer2024quality,foo2024disentangling}.

To facilitate understanding of Singlish, \cite{chow2024word} constructed a Singlish dictionary aided by ChatGPT. \citet{wang2023seaeval} built and evaluated a multicultural understanding model for the Singlish language, and performed a Singlish-to-English translation task. \citet{liu2022singlish} attempted machine translation with neural architectures such as the BART model, with decent success (METEOR score of 81.6). \citet{foo2024disentangling} attempted Singlish-to-English translation of Reddit texts using a GPT-2 LLM, and identified that the machine translation were inaccurate.

\begin{figure*}
\centering
\includegraphics[width=0.85\textwidth]{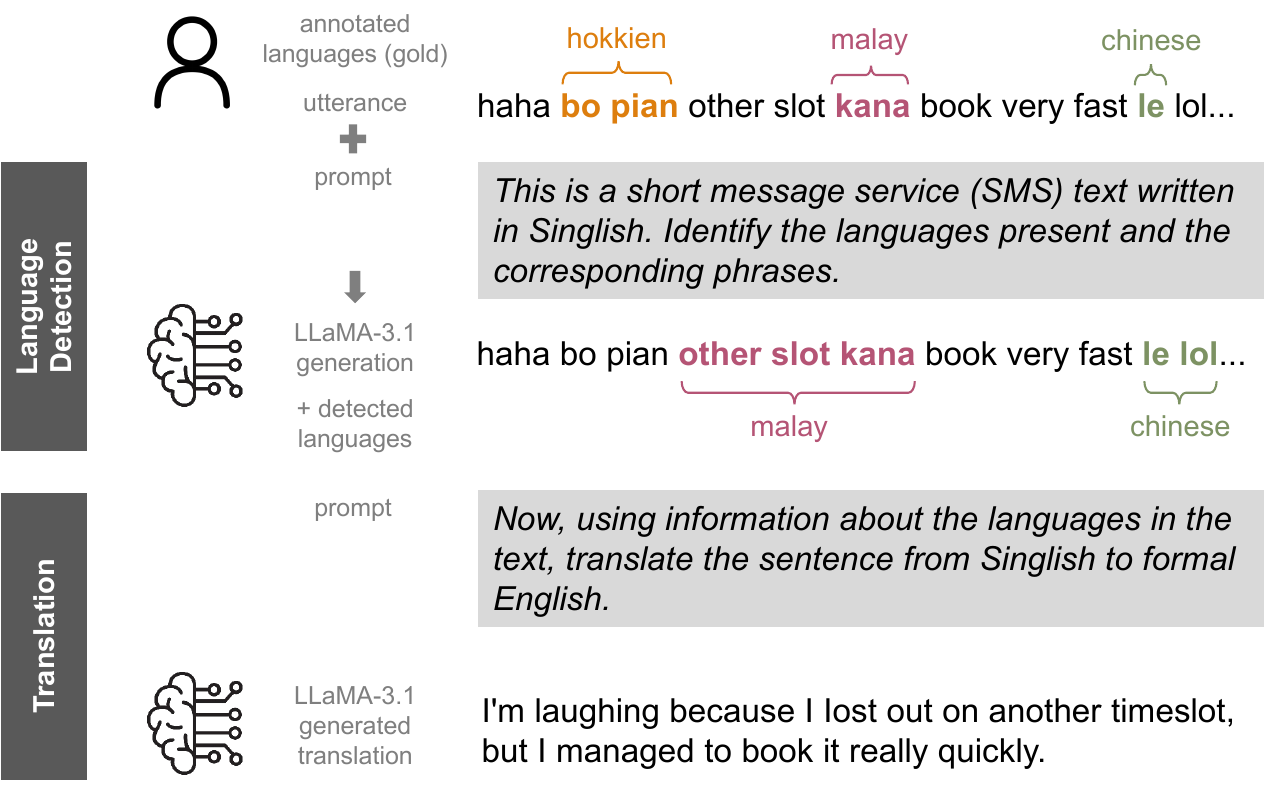}
\caption{We investigate a Language Detection and a Sentence Translation task using 5 LLMs. We prompt the LLMs using a multi-step prompting scheme, where prompts are illustrated in the grey boxes.}
\label{fig:diagram}
\end{figure*}

\section{Preliminaries}
What makes a code-mixed sentence? Apart from comprising vocabulary from various languages and language variants, a code-mixed sentence can also assume grammar structure of its superstrate languages while only comprising English terms. As a result, non-Singlish speakers might find it difficult to fully understand a Singlish sentence even if it only comprises English words. This Singlish ``what talking you" loosely translates to ``what are you talking about", and typically conveys feelings of frustration and disdain. 

In this work, we focus on other-language artifacts in code-mixed Singlish. 

\section{Task Definitions \& Methodology}
We perform two tasks with code-mixed Singlish texts. 

\begin{enumerate}
    \item \texttt{\textbf{Language Detection}}. Given a Singlish sentence as input, the output should be the languages present in the text and the corresponding phrases that use that language. Canonically, the language detection task infers the language of an entire sequence from its given context. In our experiments, we aim to detect the all the languages and language variants present in a sequence, which can be more than one. We aim to detect non-English vocabulary, which we refer to as other-language.
    \item \texttt{\textbf{Translation}}. Given a Singlish sentence as input, the output should be a sentence translated to standard formal English. We loosely refer to standard English as the form of the language that can be widely understood by the English-speaking population, regardless of nationality or culture. This is the English we interface with on news sites, or government websites. 
\end{enumerate}

We describe the formation of \texttt{Reference Texts} through expert annotators, and \texttt{Generated Texts} through LLMs as automated annotators in the following subsections.

\subsection{Source Texts}
Our dataset is a subset of the \texttt{NUS-SMS-Corpus} \cite{NUS-SMS-Corpus}, which contains texts from Short Messaging Services (SMS). The texts from this corpus demonstrates how Singlish is used in daily conversation. We randomly sampled 300 sentences with word length > 20 from the corpus to form the \texttt{\textbf{Source Texts}}.

\subsection{Reference Texts}
Three native Singlish speakers were recruited for the formation of the Reference Texts corpus. All speakers had at least a Bachelors' education in a university with English as the medium of instruction. For the language detection task, the speakers annotated the portions of the texts that were written in languages other than English. These languages and variants variants include: Chinese, Malay, Indonesian, Hokkien and Teochew. Hokkien and Teochew are dialects of Chinese. For the translation task, the annotators translated the Singlish sentence to formal English with the least amount of changes possible. They did not paraphrase unless it is necessary to express the same meaning in standard and grammatical English. Both translations and language annotations were cross-checked between the annotators. 

\subsection{Generated Texts}
We used 5 open-sourced pre-trained LLMs to perform both language detection and translation tasks. We choose to use LLMs because these models can handle multilingual sentences, and are thus suitable for Singlish sentences that contain spans of multiple languages. These models are: Mistral-7B-Instruct-v0.3-Q4\_K\_M.gguf \cite{jiang2023mistral}, meta-llama-3.1-8b-instruct \cite{llama3modelcard}, gemma-2-2b-instruct \cite{team2024gemma}, qwen2.5-14b-instruct \cite{peng2023yarn}, phi-3.1-mini-128k-instruct \cite{abdin2024phi}. The models are obtained from the HuggingFace library\footnote{\url{https://huggingface.co/}}. In the rest of the paper, we refer to these models as: Mistral-7B, LLaMA-3.1, Gemma-2, Qwen-2.5, and Phi-3.1 respectively.

For each text, we use a multi-step prompting scheme as illustrated in Figure \ref{fig:diagram}. Language detection is first conducted, where the prompt asks the LLM to identify the languages present in the text and its corresponding spans from the source text. The next step incorporates the output from the previous step, and prompts the LLM to translate the sentence from Singlish to formal English. All prompts were carried out at a temperature of 0.7. Verbalizers used can be found in Appendix \ref{sec:prompts}.

\section{Results}
In this section, we conduct quantitative analysis on the performance of language detection and translation tasks. We also present some qualitative examples to illustrate the results.

\subsection{Language Detection}
We processed the output of the Language Detection step that were returned by the LLMs. We compared the predicted languages by the LLM against the annotated set of actual languages and normalized the results against the total number of actual languages identified.
We also noted down the most common predicted languages that were not within our predefined list of language labels, such as Hindi and Singlish. Note that we ignored sentences where the LLMs did not return a result, or returned a malformed result. 

\paragraph{Quantitative Evaluation}

\begin{figure*}
\centering
\includegraphics[width=\linewidth]{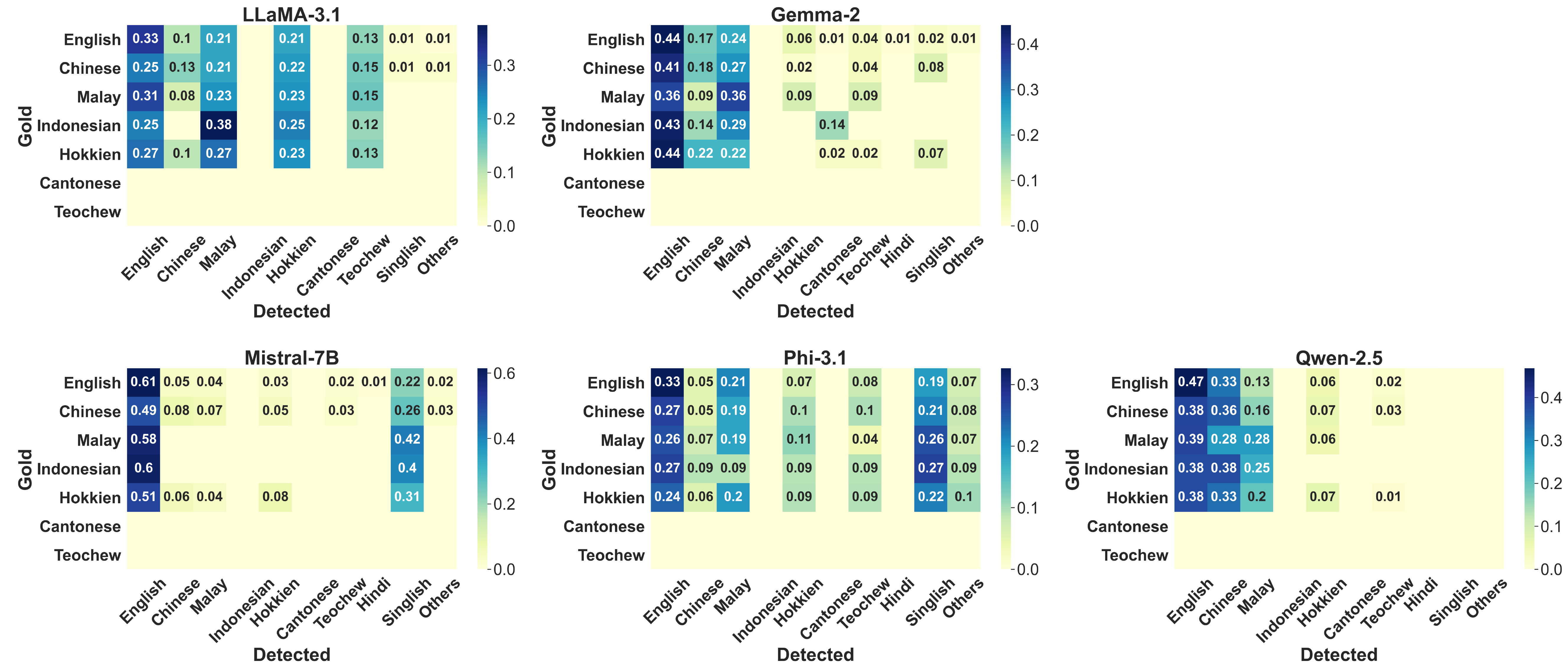}
\caption{Confusion matrix presenting predicted / detected language labels by LLMs, against gold language labels. We unit normalize each row for comparison across models. }
\label{fig:language_evaluation}
\end{figure*}

Figure \ref{fig:language_evaluation} presents correlation plots of the actual languages as reflected in predicted languages by LLMs. The Qwen-2.5 model returned the most number of valid responses, while LLaMA-3.1 returned the least number of valid responses. Surprisingly, LLaMA-3.1, Mistral-7B and Phi-3.1 are able to correctly identify that the sentences were written in Singlish, with Phi-3.1 expressing this capability to the highest extent.

All models except LLaMA-3.1 are able to identify languages that were not in the predefined set of language labels (English, Chinese, Malay, Indonesian, Hokkien, Cantonese and Teochew), indicating that the LLMs are able to draw from intrinsic knowledge. Interestingly, all LLMs (apart from LLaMA-3.1) identified the Hindi language as part of some of the texts. We observe that Hindi is one of the languages that is constantly code-mixed within Singlish. This finding provides hope that LLMs are able to cope with the evolution of language usages, particularly in the case of a language like Singlish, where Asian influences are constantly being incorporated into the Singlish vernacular. 


\paragraph{Error Analysis.}
To provide a comprehensive analysis of the language detection module, we identify 3 main types errors, and provide examples for each error type in Table \ref{table:ld_error}.
\begin{enumerate}
    \item Misidentified span but correct label. 
    \item Correctly identified span but incorrect language prediction. 
    \item Identified span that is English (false positive). 
\end{enumerate}

\begin{table*}
    \centering
    \begin{tabular}{c c | p{3cm} p{2.5cm} | c c}
    \hline
       \textbf{Error Type} & \textbf{Model}  & \textbf{Source} & \textbf{Identified Span} & \textbf{Predicted Label} & \textbf{Gold Label} \\ \hline 

    (1) & LLaMA-3.1 & ...other slot \textbf{\textit{kana}}... & other slot kana & Malay & Malay \\

    (2) & Mistral-7B & \multicolumn{2}{|c|}{\textbf{\textit{matter pata nai tha}}} & Malay & Indo \\

    (2) & LLaMA-3.1 & \multicolumn{2}{|c|}{\textbf{\textit{nait kai siu hei chong lai}}} & Teochew & Hokkien \\ 

    (3) & Phi-3.5 & \multicolumn{2}{|c|}{wil nevr go unrecognized} & Malay & N.A \\

    \hline
    \end{tabular}
    \caption{Examples of errors identified. Actual (gold) span indicated in bold italics.}
    \label{table:ld_error}
\end{table*}


Error type (1) represents span-detection errors but with accurate language detection. LLaMA-3.1 incorrectly identifies preceding words as also being Malay even though that is not the case. Empirically, this error is observed more frequently when the source text's structure is different from that of standard English. We posit that this out-of-distribution text contributes to this error. More robust analysis have to be conducted to draw any firm conclusions. We note here that while the intermediate span-detection task fails, the main language detection task is correct. 

Next, error type (2) is inaccurate label detection. There are also examples where the language is wrongly identified, but the span is correctly highlighted. Mistral-7B correctly identifies the span ``\textit{matter pata nai tha}", but it predicts the span to be Malay instead of Indonesian. Similarly, LLaMA-3.1 identifies the span ``\textit{nait kai siu hei chong lai}" as Teochew instead of Hokkien. These two examples show regional languages and dialects are similar to one another; LLMs are unable to differentiate such differences and misclassify these labels. 



Lastly, error type (3) includes errors where the LLM identifies a span where there is no other-language detected. The Phi-3.5 model identifies the phrase ``wil nevr go unrecognized" as in the Malay language, likely due to the abbrievation (or misspelling) of the word ``will" as ``wil" and ``nevr" to ``never". 

From our results, LLMs can aid in the Language Detection task, though they are lacking in accuracy for more nuanced languages.

We also curate a list of phrases that are generally incorporated within the Singlish texts and the languages they are adapted from in Section \ref{sec:lookup}. This list provides context to the languages that Singlish adapts from. 

\subsection{Sentence Translation}

We evaluate the generated translations against the reference translations written by human annotators using BLEU, ROUGE and BertScore.

BLEU \cite{bleu} and ROUGE \cite{rouge} capture lexical similarity between source and generated text, by taking into account the precison and recall of n-gram overlaps. We report ROUGE-L, which is based on the longest common subsequence overlap. BertScore aims to capture semantic similarity by capitalizing on sequence context and is reported to correlate relatively well with human judgement \cite{bertscore}. We report BertScore-F1.

\paragraph{General Evaluation.}

\begin{table}[h]
\centering
\begin{tabular}{l|c|c|c}
\hline
 \textbf{Model}   & \textbf{BLEU} & \textbf{ROUGE} & \textbf{BertScore}  \\\hline
baseline      & \textbf{0.4358} & \textbf{0.6701} & 0.9253 \\
Phi-3.1       & 0.1297 & 0.2980 & 0.8882 \\
LLaMA-3.1     & 0.2303 & 0.4531 & 0.9110 \\
Gemma-2       & 0.3630 & 0.6056 & \textbf{0.9312} \\
Qwen-2.5      & 0.3003 & 0.5908 & 0.9269 \\
Mistral-7B    & 0.2795 & 0.5295 & 0.9194 \\
\hline
\end{tabular}
\caption{Performance of LLMs on rewriting Singlish texts to formal English. Best performance in bold.}
\label{table:mt-perf} 
\end{table} 

\begin{figure}[h]
\centering
\includegraphics[width=\linewidth]{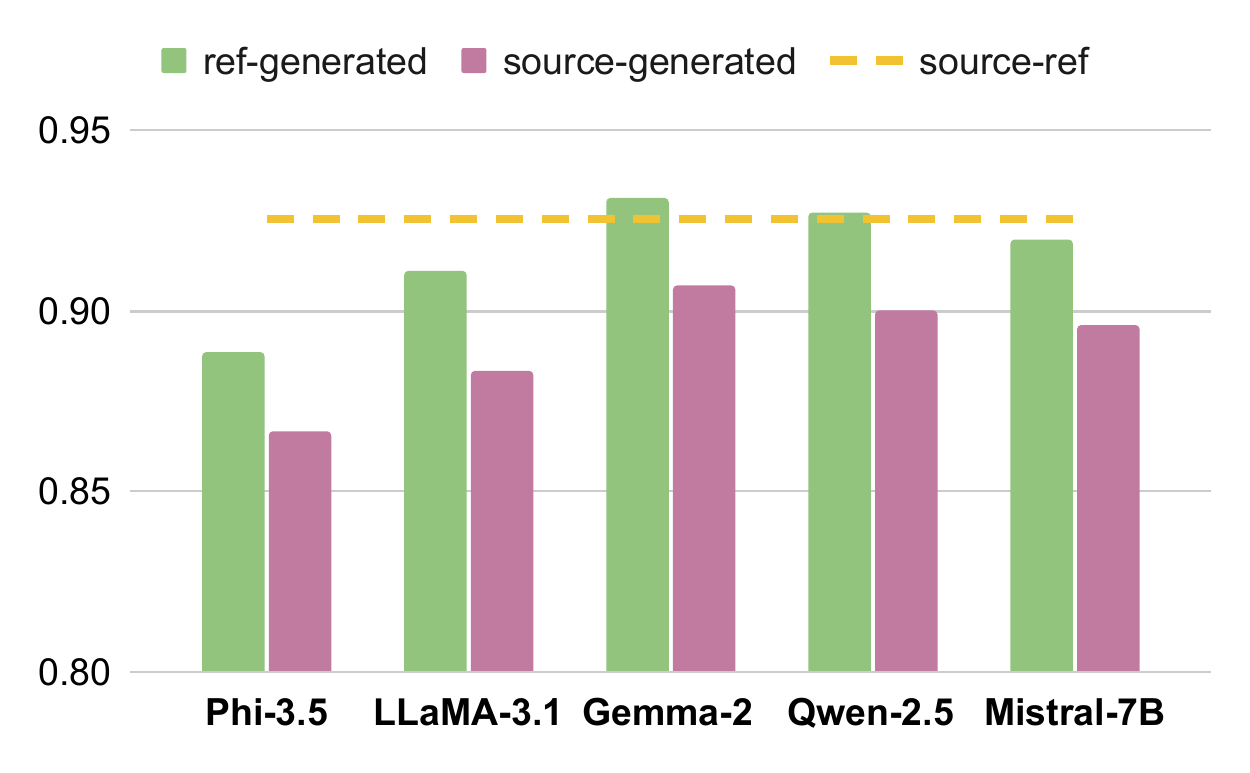}
\caption{BertScore performance of generated text against source text. Dotted yellow line represents the theoretical oracle performance when source text is evaluated against human reference text.}
\label{fig:compare_src}
\end{figure}

For the sake of a reference in comparison, we include the baseline oracle performance when source text is used as reference, and human-translated references are used as candidate texts. Table \ref{table:mt-perf} shows that the performance on the translation task varies across the different LLMs. The variance of performance across the different LLMs necessitates the search of the ``most suitable" model for this task. 

While Gemma-2 performs the best in lexical matching metrics, there still exists a large gap between its generations and the oracle baseline performance. Interestingly, we see a high semantic similarity between Gemma-2's generations. This suggests that while the generations might be paraphrased, contributing to relatively lower lexical similarity scores, the meaning of the text is largely retained. 

Figure \ref{fig:compare_src} shows the difference in BertScore performance of generated text against source and reference texts respectively. The dotted yellow line represents the theoretical upper-bound for source-generated metrics only, as both metrics use source texts as the reference text. We see that there exists a gap between reference-generated and source-generated translations across all models. This suggests that there is still room for improvement for generation translations to close the semantic gap against source texts. 

The authors note that the oracle BertScore is lower than that of Gemma-2. While we did not claim that the oracle score is the theoretical upper-bound for reference-generation evaluation, we clarify that the lower oracle score is possibly due to the fact that the source text (Singlish) contains many non-English tokens that are unknown to the underlying evalution model. This is an intrinsic limitation that we do not address in this work. 

\begin{figure*}%
    \centering
    \subfloat[\centering BLEU]{{\includegraphics[width=0.45\linewidth]{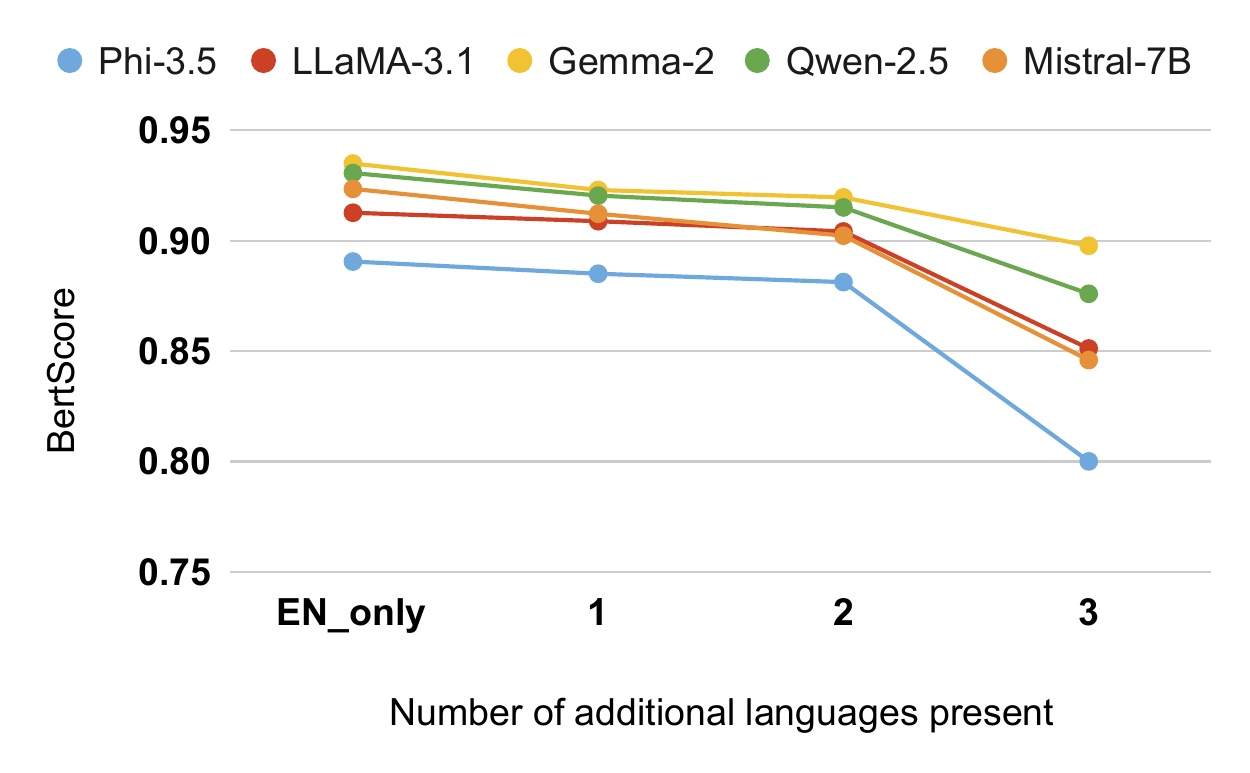} }
    \label{fig:bleu_numlang}}%
    \qquad
    \subfloat[\centering BertScore]{{\includegraphics[width=0.45\linewidth]{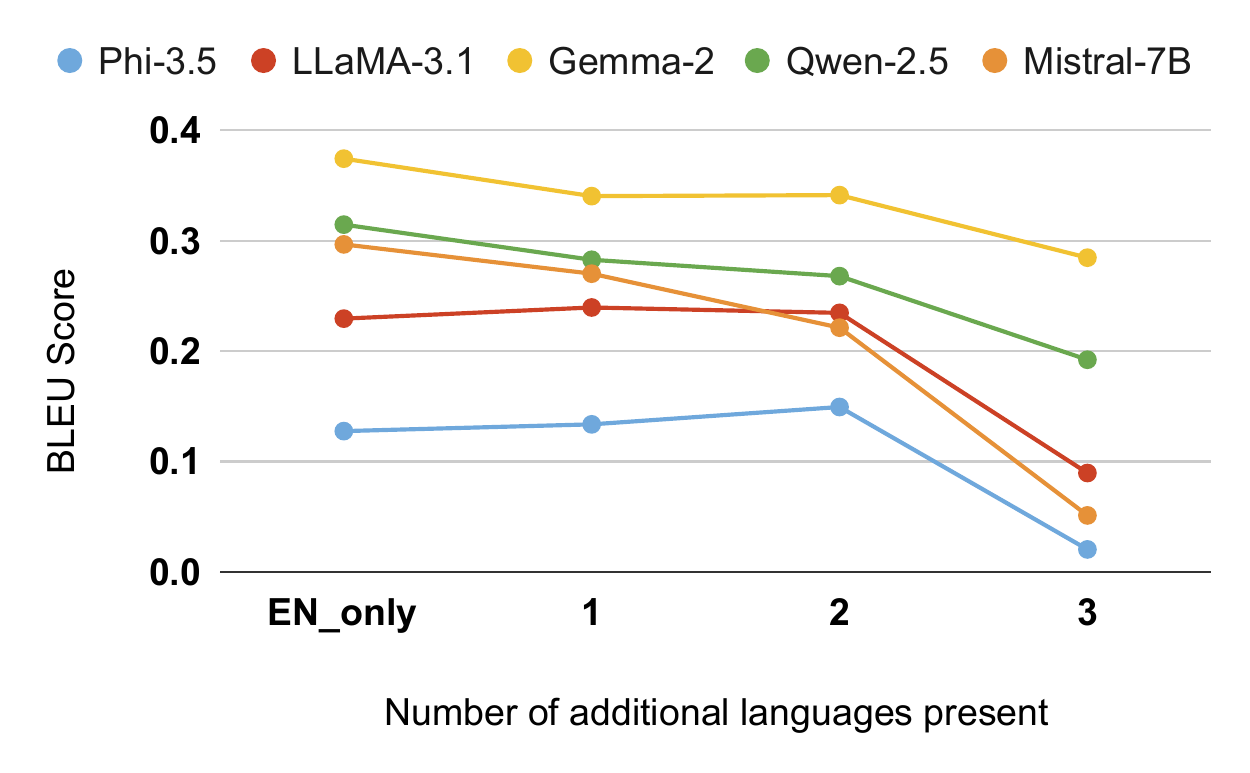} }
    \label{fig:bs_numlang}
    }%
    \caption{Performance of LLM translation against the number of languages present in source text.}
    \label{fig:numlang}%
\end{figure*}

\paragraph{Performance Degradation over Number of Languages Present.}
An interesting dimension of analysis is how the degree of code-mixing affects translation performance. We use number of additional languages (on top of English) present as a proxy for the degree of code-mixing in a sentence. Figure \ref{fig:numlang} shows that translation performance degrades as more languages are present in the source text across both lexical and semantic metrics. The dip in performance is especially notable when more than 2 additional languages are present. This observation is within expectations: if a sentence contains only English tokens, then the machine requires less effort to re-write the sentence into formal English structure. A well-trained LLM is assumed to be able to generate fluent text, which we expect when we output ``formal" English texts. Conversely, if a sentence contains multi-lingual tokens, then the machine requires more effort to understand and translated these tokens before generation. 

\paragraph{Limitations in translation.}
As reiterated throughout the paper, Singlish relies heavily on context for accurate understanding. As a result, machine translations can seem correct, but is actually wrong. 

\begin{table}[h]
\centering
\begin{tabular}{l p{5.5cm}}
\hline
 \textbf{source}   &  haha bo pian other slot kana book very fast le lol\\ \hline
 \textbf{generated} & I'm laughing because I lost out on another time slot, but I managed to book it really quickly. \\ 
 \hline
 \textbf{reference} & I did not have a choice in choosing this slot, because the other slots were booked out very quickly. \\
\hline
\end{tabular}
\caption{Generations by LLaMA-3.1.}
\label{table:mt-qual} 
\end{table} 

Table \ref{table:mt-qual} shows an example of how the generated translation may seem to be reasonable, but has a different meaning to the intended message. The generated text suggests that the subject managed to ``book a slot" quickly. However, the intended meaning of the sentence is that the subject was unable to obtain his desired slot, as ``the other slots were booked out quickly". This example highlights the nuances in code-mixed languages, and why they are difficult for machines to interpret and translate. We further expound challenges in translating code-mixed texts in Section \ref{sec:challenges}.

Last, as with all LLM generations, our translations also suffer from hallucinations, where information not appearing in the source text appears in the generation. We provide some examples in Section \ref{sec:mt-examples}. We leave it to future work to circumvent such hallucinations.

\section{Challenges in Translating Code-Mixed Texts}
\label{sec:challenges}
Our results show that language models are not yet proficient in translating code-mixed sentences. In this section, we elaborate on some challenges in translating code-mixed texts to English, and specifically in translating Singlish texts to English. 

\paragraph{Lack of datasets.} There are limited datasets for code-mixed languages that are translated by experts into formal English. This dearth of resources limits the advancements that researchers can make in constructing accurate models. Examples of code-mixed datasets include: a Malayalam-English dataset \cite{chakravarthi2020sentiment}, a Hindi-English dataset \cite{nayak2022l3cube}, a Hindi-English and Bengali-English dataset \cite{patra2018sentiment}, and so forth.

However, these datasets contain only two languages in code-mixed sentences. To that end, we release our manually translated dataset, which contained code-mixed sentences in multiple languages. We hope to facilitate machine translation for texts code-mixed with multiple Asian languages. To the best of our knowledge, this is the first dataset in Singlish that is annotated for code-mixed phrases, with translations from Singlish to English.

\paragraph{Sentence structures formed from derived languages.} For code-mixed sentences, the parts of the sentences that are not in English typically adopts the language structure of its borrowed language. Therefore, typical NLP tools for sentence processing in English will not be able to accurately process the sentence. A legitimate Singlish sentence would be ``i just now go down eat", which does not have a legitimate English sentence structure. However, this sentence structure is the original structure of its derived language, Chinese, \begin{CJK*}{UTF8}{gbsn}我刚才下去吃\end{CJK*}, and the Singlish sentence is a direct translation of the Chinese phrase.

\subsection{Challenges in Singlish Translation}
Singlish is a complex language which code-mixes multiple Asian languages. Our experiments show that current state-of-the-art language models are still lacking in the Language Detection and translation of Singlish texts. Here, we make a few observations of the challenges in Singlish-to-English translations in the following section. These translation problems are specific to the Singlish language.

\paragraph{Lack of subject.}
The apparent lack of subject in Singlish sentences makes it challenging for machines to determine the speaker, or the topic of the sentence. While native speakers can infer from preceding conversational contexts, tone and attitude of a speaker, such information is unavailable to LLMs. Appendix \ref{sec:mt-examples} presents examples of how the LLMs misinterprets the subject; in the sentence ``surely bring us tour around the school one", all LLMs either infers tour-giver to be the speaker or the receiver when the speaker refers to both himself and his audience as tour participants. That is, the tour-giver is a third party.

\paragraph{Ambiguity of tenses.}
Most often, Singlish typically uses the present tense, bearing resemblance to the Chinese and Malay languages where verbs are not inflected for tenses \cite{lin2006time,svalberg1998english}. For example, \begin{CJK*}{UTF8}{gbsn}我下去吃\end{CJK*} translates to the Singlish sentence ``I go down eat". This sentence has no indication whether the author meant that he is presently heading down to eat, or had already head down to eat. Other times, to express an action in the past, Singlish texts intentionally suffix a word with ``-ed". An instance from our dataset is ``solo-ed", to indicate that the author had previously performed the action alone. Both interpretations are valid, and thus contextual cues from neighboring sentences are required to better understand the sentence. Therefore, an added challenge in Singlish translation is diambiguating the tenses within a sentence, to properly reflect the author's intentions.

\paragraph{Singlish drops the transitive verbs.} One of the Singlish sentences from the dataset is: ``Okay. By the way, can I ask: when do you need the 2 people? Is it for the same exhibition on this coming weekend?". A translated version of this sentence would contain several transitive verbs that were dropped in the construction of the Singlish sentence: ``Okay. By the way, can I ask: when do you need the 2 people? Is it for the same exhibition [happening] on this coming weekend?" Singlish is a language that embodies conversational brevity \cite{lai2023singlish}, where we observe missing transitive verbs. A language model thus has to infer these phrases from the implicit information embedded in the sentence or its surrounding sentences. 

\paragraph{Singlish drops the mid-sentence punctuations.} 
A Singlish sentence may sometimes look like a run-on English sentence, particularly because of the lack of mid-sentence punctuations. For example, ``i just solo-ed church ruins hard with my friends watching haha then suddenly one hobgoblin appear behind me i got shock then unglam moment when i was like omg hahaha", is written entirely in English, but requires some thought for better understanding. Singlish compresses multiple sentences syntactically by removing unnecessary parts of a message, which consequently means that punctuations are typically optional \cite{tem2022syntactic}. Some ways past research fixed this property is to insert the punctuation where necessary \cite{liu2022singlish}. Singlish translation tasks therefore requires a level of natural language understanding in order to add these missing punctuations.

\section{Limitations and Future Directions}
We acknowledge several limitations within our work. We tested the language capabilities on open-sourced LLMs only, and the use of only one prompting scheme due to limited resources. Subsequently, further investigation can be done with different types of prompting schemes, such as zero-shot prompting or Chain-Of-Thought prompting, which may increase accuracy of the translation task. In deploying our LLMs, we used the pre-trained model. For better understanding of the language, further work calls for a fine-tuned model specific for the Singlish language. Secondly, our dataset is primarily based on the Short Message Service (SMS), which can be very different from the Singlish spoken orally, or the Singlish written on the web, which therefore warrants opens future studies into translation of such colloquial forms of Singlish. Lastly, further extensions of this work can include investigating why models identify phrases as a particular language, for example, because models evaluate sentence structure or similar words.

Despite the challenges in translating code-mixed languages, there are several future directions that Singlish translation work could take.

\paragraph{Understanding of usage of code-mixed languages in informal settings.} Code-mixing is generally used for social communication such as on social media or on forums \cite{santy2021bertologicomix}. An understanding of code-mixed sentences thus provides a window into online opinions, which are especially important in crucial events such as political elections and natural crisis. Many of such analysis only examine English-only texts, thus missing out the important opinions of native people, so conclusions can be ill-informed. As an example, \citet{jacobs2023tracking} and \citet{ng2024tiny} examined Asian political events in Taiwan and Indonesia, but only the English-based tweets, leaving out crucial information in the tweets written in the code-mixed languages of Chinese-English and Indonesian-English. Downstream work includes large-scale translation of such code-mixed posts from social media and investigating the opinion dynamics.

\paragraph{Understanding code-mixed languages through contextual cues.} The interpretation of code-mixed languages must take into account cultural cues from the language and contextual cues from the surrounding sentences. Such cues can come in the form culture-specific and language-specific norms, and implied messages from surrounding sentences. A machine translation model must be fine-tuned with such nuances, which involves deeper research into the semantic and pragmatic space of language understanding \cite{santy2021bertologicomix}.

\section{Conclusion}
We examined the translation capabilities of LLMs for code-mixed languages. We observe that while LLMs do not always accurately predict the languages used in a sentence correctly, they are able to identify other Asian languages present in the sentence, and are mostly able to identify that the language is in Singlish. In sentence translation, while LLMs can translate a Singlish sentence to a grammatically correct English sentence, the translation disregards context, and is thus an incorrect translation.

Our investigations show that there is hope for LLMs to be used for understanding code-mixed languages, especially Singlish. Our work introduces a Singlish code-mixed dataset, annotated for the presence of Asian languages, and translated into formal English.  We detail several challenges that are faced in the translation task of code-mixed languages, and present potential opportunities for the computational study of such languages. We encourage more native speakers to actively engage in this area of NLP research.

\section*{Acknowledgments}
The authors thank Russell Goh for his help with formal English translations and language annotations. The authors thank the National University of Singapore Development Grant for funding the work.

\bibliography{custom}

\clearpage
\appendix

\section{Appendix}
\subsection{List of Singlish words derived from Asian Languages}
\label{sec:lookup}
Table \ref{tab:lookup} lists words and phrases derived from Asian languages that are found in the Singlish texts.

\begin{table}[H]
    \centering
    \begin{tabular}{rp{5cm}}
    \hline
       \textbf{Language}  & \textbf{Phrase} \\ \hline 
       Chinese & liao, le, lorh/ lor, ma, mah, de, wah, hai/ haiya/ haiz, ah, eh, err/ er, ba/ bah \\ 
       Hokkien & jio, paiseh/ ps, chop chop, huh/ ha, bo bian/ bui, heng, sia \\
       Malay & shiok, kana \\
    \hline
    \end{tabular}
    \caption{Phrases commonly found in Singlish texts and the language they are adapted from}
    \label{tab:lookup}
\end{table}

\section{Prompts used for Large Language Models}
\label{sec:prompts}
We performed a multi-step prompting scheme with Large Language Models for the Language Detection and the Sentence Translation tasks. Here are the prompts used: 

\paragraph{For Language Detection}
The language detection prompt takes in the message as a parameter and asks the LLM to identify the languages used in the short text messages written in Singlish.

\begin{verbatim}
prompt = f"This is a short message service 
(SMS) text written in Singlish. Identify 
the languages present and the 
corresponding phrases. 
Message: {message}"

messages=[
    {"role": "assistant", 
     "content": """
         You are an AI that identifies 
         languages used in Singlish messages. 
         Identify the languages and 
         corresponding phrases. 
         Return an array consisting of 
         [{"language": language, "phrase": phrase}] 
         in valid JSON format. 
         Do not return any other text.
     """},
    {"role": "user", "content": prompt}
]
    
\end{verbatim}

\newpage
\paragraph{For Sentence Translation}
The sentence translation prompt not only takes the text message as a parameter, but also takes in the information about the languages detected from the previous step as a parameter. The language information helps the LLM better understand the Singlish sentence.

\begin{verbatim}
prompt = f"Now, using information about 
the languages in the text: {language_info}, 
translate the sentence from Singlish 
to formal English. 
Message: {message}"

messages=[
    {"role": "assistant", 
     "content": """
         You are an AI that translates Singlish 
         to formal English. Use the 
         identified languages and phrases 
         to provide an accurate translation. 
         Return only the translated text, 
         and no other text.
     """},
    {"role": "user", "content": prompt}
]    
\end{verbatim}

\section{Example of Generated Translations}
\label{sec:mt-examples}
Table \ref{table:mt-gen} presents a sample of the generated translations from each of the Large Language Models.

\begin{table*}[h]
\centering
\begin{tabular}{l p{14.5cm}}
\hline
 \textbf{model}   &   \textbf{text} \\ \hline

 source & hey xin,mon 4 aug, 0845 at lt32,formal wear, cant wear jeans.tats d plan for 1st day of sch.haiz, surely bring us tour ard e sch one, sianz... \\ \hline

 reference & Hey xin, on monday 5th of August, we are to meet at 0845 at LT32. Attire is formal wear, no jeans. That's the plan for the first day of school. I'm sure they will bring us to tour the school, sigh. \\ \hline
 
 Phi-3.5 & Subject: Formal Attire Requested For School Trip on August 4th - Mr./Ms. [Your Name]

Dear Sir/Madam,

I am writing to inform you that I will be attending a school trip scheduled for the first day of classes in late August (August 3rd or 4th). It is imperative that my outfit consists solely of formal wear. Please ensure that jeans are not included as part of the appropriate dress code, and kindly consider this request while organizing any clothing-related provisions for me.

I am particularly interested in extending an invitation to join a special school tour on August 4th at LT32 (or whichever location is relevant). It would be greatly appreciated if you could arrange transportation and accommodations necessary for this unique opportunity, as well as provide any required information or guidance.

Thank you in advance for your cooperation. I look forward to the event with great anticipation!

Kind regards,
[Your Name]\\ \hline

LLaMA-3.1 & Hello Xin, Monday (4th August) at 8:45 am at LT32, formal attire is required and jeans are not allowed. That's the plan for our first day of school. Haiz, I'll definitely give you a tour around the school, okay?
 \\ \hline
Gemma-2 & Hi Xin,

Meet on August 4th at 8:45 AM in front of LT32.  Formal attire is required; no jeans allowed. That's the plan for the first day of school. Hopefully you can bring us around the school. Sigh, I know it's going to be a long day.
\\ \hline
Qwen-2.5 & Hello Xin, on Monday, August 4th, at 0845 hours in assembly hall LT32, formal attire is required; jeans are not permitted. This is the plan for the first day of school. Sigh, be sure to give us a tour around the school, it's such a hassle...
\\ \hline
Mistral-7B & On Monday, August 4th, at 8:45 am at LT32, formal wear is required (jeans are not allowed). This is the plan for the first day of school. Please make sure to bring us your admission ticket to school, I hope. Sigh...\\

\hline
\end{tabular}
\caption{Example of Generated Translations with Multi-Step Prompt.}
\label{table:mt-gen} 
\end{table*} 

\end{document}